\documentclass[letterpaper, 10 pt, conference]{ieeeconf}  

\IEEEoverridecommandlockouts                              
\usepackage{graphicx}
\usepackage{mathtools}
\usepackage{amssymb}
\usepackage{algorithm, algorithmicx, algpseudocode}
\usepackage{epstopdf}
\usepackage{xcolor}
\usepackage{hyperref}
\usepackage{adjustbox} 
\usepackage{caption} 
\usepackage{floatpag}
\usepackage{subcaption}
\usepackage{graphicx}
\usepackage{multirow}
\usepackage{booktabs}
\floatpagestyle{empty}

\title{\LARGE \bf
GluMind: Multimodal Parallel Attention and Knowledge Retention for Robust Cross-Population Blood Glucose Forecasting 
}

\author{Ebrahim Farahmand$^{\star}$, Reza Rahimi Azghan$^{\star}$,  Nooshin Taheri Chatrudi$^{\star}$, Velarie Yaa Ansu-Baidoo$^{\ddagger}$, \\ Eric Kim$^{\star}$, Gautham Krishna Gudur$^{\dagger}$, Mohit Malu$^{\star}$, Owen Krueger$^{\star}$, \\Edison Thomaz$^{\dagger}$, Giulia Pedrielli$^{\star}$, Pavan Turaga$^{\star}$, Hassan Ghasemzadeh$^{\star}$
\thanks{$^{\star}$Arizona State University, Phoenix, AZ, USA\vfill$^{\ddagger}$ University of Miami, Miami, FL, USA \vfill$^{\dagger}$ The University of Texas at Austin, Austin, TX, USA}%
}

\begin{document}

\maketitle
\thispagestyle{empty}
\pagestyle{empty}

\begin{abstract}
Diabetes is a chronic metabolic condition marked by consistently elevated blood glucose levels (BGLs), which can lead to serious adverse health outcomes such as cardiovascular disease, stroke, and neuropathy. Accurate forecasting of BGLs helps individuals maintain normal glucose levels and enables caregivers to implement timely lifestyle interventions. However, forecasting BGLs remains a complex challenge because blood glucose is severely influenced by lifestyle factors such as physical activity, diet, and sleep. Furthermore, a person's blood glucose response to lifestyle factors highly depends on the person's insulin resistance status, resulting in significantly different glucose levels among healthy individuals, people with prediabetes, and patients with diabetes. Despite advancements in deep learning, predicting long-term BGL using multimodal and irregularly sampled sensor data remains an active area of research. Moreover, model fine-tuning across different cohorts in blood glucose predicting models leads to the catastrophic forgetting problem in deep learning-based forecasting models. To address these challenges, we propose GluMind, a transformer-based multimodal framework designed for continual and long-term blood glucose forecasting. GluMind devises two attention mechanisms, including cross-attention and multi-scale attention, which operate in parallel and deliver an accurate predictive performance. Cross attention effectively integrates blood glucose data with other physiological and behavioral signals such as activity, stress, and heart rate, addressing challenges associated with varying sampling rates and their adverse impacts on robust prediction. Moreover, the multi-scale attention mechanism captures long-range temporal dependencies. To mitigate catastrophic forgetting, GluMind incorporates a knowledge retention technique into the transformer-based forecasting model. The knowledge retention module not only enhances the model’s ability to retain prior knowledge but also boosts its overall forecasting performance. We evaluate GluMind on the recently released AIREADI dataset, which contains behavioral and physiological data collected with healthy people, individuals with prediabetic, and those with type 2 diabetes. We examine the performance stability and adaptability of GluMind in learning continuously as new patient cohorts are introduced. Experimental results show that GluMind consistently outperforms other state-of-the-art forecasting models, achieving approximately 15\% and 9\% improvements in root-mean-squared error (RMSE) and mean-absolute-error (MAE), respectively.

\end{abstract}
\section{Introduction}
Diabetes mellitus is a chronic metabolic disorder characterized by high blood glucose levels resulting from the pancreas not producing sufficient insulin or failure of the body to effectively utilize the insulin that is produced~\cite{tayek2018importance}. Diabetes can be classified as type 1, type 2, and gestational diabetes. Of these three, type 2 diabetes is the most common type of diabetes affecting over 90\% of reported cases. In type 2 diabetes, there is total or partial insulin insufficiency as a result of $\beta$-cell dysfunction. Uncontrolled diabetes leads to complications including macrovascular conditions such as cardiovascular diseases and microvascular conditions such as diabetic retinopathy, neuropathy, and nephropathy~\cite{beckman2016vascular}. As of 2021, diabetes was the cause of over 1.6 million deaths; it has been projected by the International Diabetes Federation that by 2045, about 1 in 8 adults ($\sim$783 million) will be living with diabetes~\cite{roth2018global}. Additionally, the increase in diabetes prevalence poses an economic burden due to higher medical costs and loss of productivity as a result of illness, disability, and early death. In the United States, the annual economic cost of diabetes in 2022 was reported as \$412.9 billion; thus, diagnosed diabetes accounted for 1 of 4 health care dollars spent~\cite{parker2024economic}. Due to these factors, new technologies that facilitate lifestyle modifications for diabetes prevention and management are needed.

Preventing and managing diabetes is essential to health, and thus, several individual-centered approaches have been recommended to achieve this goal. Continuous glucose monitoring (CGM) devices are commercially available technologies used in the treatment of diabetes to measure glucose within the interstitial fluid. It has the capability to measure glucose in near real-time (every 5 minutes) and shows the variation in measurements. Additionally, the frequency of measurements allows for the examination of trends in blood glucose levels in an individual, helping physicians make therapeutic decisions~\cite{vazeou2011continuous}. Until recently, the use of CGMs was limited to patients with diabetes. However, with the advent of over-the-counter CGMs, these technologies are becoming pervasive and available to general public. This provides a unique opportunity for the prevention and management of type 2 diabetes. Importantly, as these devices become widely available, we need to develop analytical tools that process sensor data in real-time and provides actionable behavioral feedback for glucose control.

Therefore, the development of signal processing and machine learning algorithms that predict blood glucose levels based on behavioral and physiological data is an important research task that will facilitate glucose control for both diabetes prevention and diabetes management ~\cite{shuvo2023deep,imrisek2022effects}. Specifically, deep learning (DL) algorithms leverage data measured from CGM devices and incorporate physiological signals such as stress levels, heart rate, and physical activity. Through the analysis of these multidimensional and complex datasets, the algorithms are capable of identifying trends and patterns in blood glucose fluctuations. This ability to capture complex dynamics enables accurate prediction of future blood glucose levels.

Sequential deep learning models, such as Recurrent Neural Networks (RNNs) including Long Short-Term Memory (LSTM) and Gated Recurrent Units (GRU), have been widely used to predict blood glucose levels in individuals with Type 1 Diabetes (T1D)~\cite{patil2024modeling}. However, these models have three major limitations: (1) they struggle to capture long-range dependencies within time-series data~\cite{zhou2021informer}; (2) they are not able to effectively handle time-series data collected at varying sampling rates; and (3) existing models do not learn continuously as the models performance on earlier tasks (i.e., patient populations) drops significantly as they are fine-tuned with data from new patients. Although LSTMs were specifically designed to address the limitation of traditional RNNs in modeling long-term dependencies, they still face challenges in capturing information over long temporal sequences~\cite{nie2022time} and in processing signals recorded with different sample rates. 

Recently, Transformer models have achieved significant success among sequential DL models in various applications such as natural language processing (NLP), computer vision, and time series forecasting. This remarkable success in time series forecasting is due to leveraging the attention mechanism, which automatically captures local and temporal dependencies within the time series data. Therefore, Transformer models have become an ideal choice for multimodal tasks involving sequential modeling. However, while the vanilla Transformer architecture~\cite{vaswani2017attention} offers valuable insight into relevant time steps for multi-horizon forecasting, it falls short in modeling the relative importance of diverse time-series features—especially those with heterogeneous sampling rates—at a given time step. Prior work aimed at improving the capability of time series forecasting through the development of Informer~\cite{zhou2021informer}. However, the introduced architecture did not achieve high accuracy in long-term predictions for time series data, especially on blood glucose prediction tasks. This limitation becomes even more prominent when leveraging sequential fine-tuning of the model in continual learning settings where training data from new patient populations become available after the training process for previous patient cohorts has concluded. These sequential models forget prior knowledge when learning a new task, leading to a significant drop in the performance of the model on previous tasks, a problem referred to as catastrophic forgetting in continual learning research.  

\textbf{Key Limitations and Associated Challenges:} We listed key limitations of state-of-the-art studies and present the associated challenges of blood glucose prediction as follows.
\begin{itemize}
    \item Difficulty in achieving accurate long-term blood glucose forecasting.\looseness=-1
    \item Mismatched temporal resolutions in data sources (e.g., CGM readings, physiological, and behavioral variables). \looseness=-1
    \item Limited clinical datasets, especially for populations with type 2 diabetes.
    \item Catastrophic forgetting during cross-cohort fine-tuning, impairing retention of previously learned patterns in blood glucose forecasting models.

\end{itemize}

\begin{figure}
    \centering
    \includegraphics[scale=0.25]{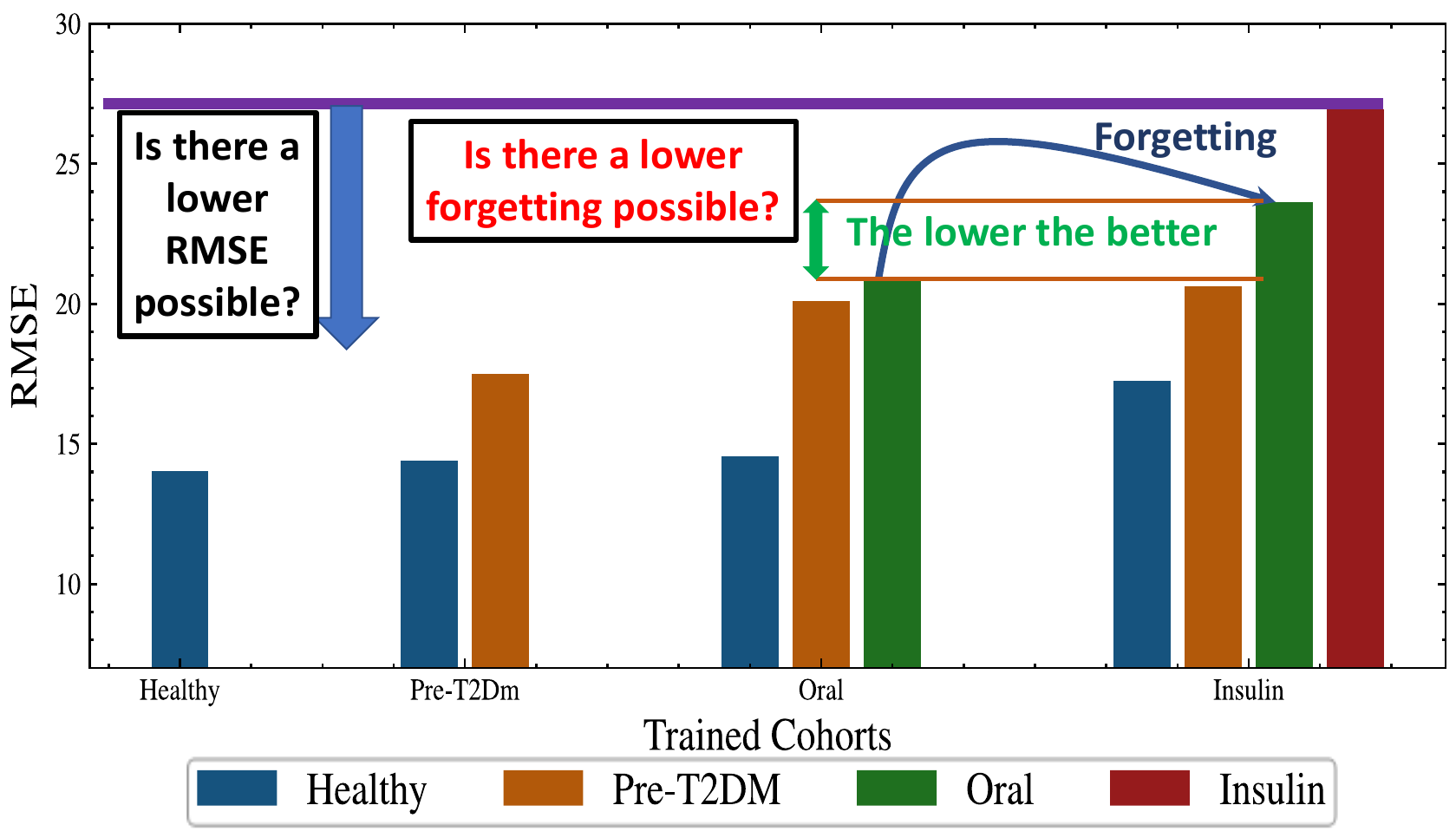}
    \caption{The Root Mean Square Error (RMSE) of the AttenGluco model~\cite{farahmand2025attengluco} for blood glucose prediction across various cohorts in the AI-READI dataset. The rising bar heights suggest that the model experiences catastrophic forgetting. These findings prompt the question of whether additional improvements in robustness and error minimization can be achieved.\looseness=-1}
    \label{fig:question}
\end{figure}

Fig.~\ref{fig:question} showcases one of the limitations of the state of the art and highlights the key research questions that we seek to address in this paper. In this figure, the AttenGluco model~\cite{farahmand2025attengluco} is applied to predict blood glucose levels across different cohorts. The model is transferred between cohorts and evaluated on individuals from the previous cohort to test its knowledge retention ability. This figure depicts that the AttenGluco model suffers from the catastrophic forgetting issue.\looseness=-1

To address these challenges, we propose a novel forecasting algorithm designed for long-term blood glucose prediction in individuals with Type 2 Diabetes (T2DM). The model builds on a Transformer architecture and incorporates two complementary attention mechanisms: cross-attention and multi-scale attention. Cross-attention captures the influence of external time series variables (e.g., physiological signals, behavioral data) on blood glucose levels. Moreover, it is well-suited for handling time-series data collected at different sampling rates. Multi-scale attention captures long-term dependencies in temporal data and handles the various temporal resolutions in data sources. Furthermore, to the best of our knowledge, this work is the first to explore the problem of blood glucose forecasting using the AI-READI dataset~\cite{AIREADI2024,AIREADI20242}, which includes CGM data and additional physiological variables from individuals with T2DM. Based on our knowledge, this study is also the first to investigate the issue of catastrophic forgetting across different cohorts in diabetes datasets and mitigate it.
The following list summarizes the novel \textit{contributions} of our work.\looseness=-1
\begin{itemize}
\item Our work introduces a transformer-based approach to BGL forecasting that uniquely combines various physiological signals for improved prediction accuracy.
\item The model's architecture features two complementary attention mechanisms working in parallel:

\begin{itemize}
    \item Cross-attention that identifies meaningful relationships between different input variables.
    \item Multi-scale attention that captures the long local semantic information throughout the input variables.
\end{itemize}

\item We tackled the personalization challenge through subject-to-subject fine-tuning, adapting the model's parameters to individual metabolic characteristics.\looseness=-1
\item To combat the observed "forgetting" effect when transferring to new patient cohorts, we integrated the Learning without Forgetting~\cite{li2017learningforgetting} (LwF) technique. This helps the model improve its knowledge retention capabilities.

\end{itemize}



\section{Related Work}
Research on blood glucose forecasting generally falls into two main categories: (i) physiologically-based models that simulate glucose dynamics using predefined equations and domain knowledge~\cite{dalla2006system}, and (ii) data-driven machine learning approaches that learn patterns directly from real-world or simulated datasets~\cite{shuvo2023deep}. While physiological models offer interpretability rooted in biological processes, data-driven models provide greater adaptability and scalability across diverse populations and conditions. This study primarily focuses on the data-driven approach. We will provide a literature review on various data-driven machine learning methods for predicting blood glucose levels.

\subsection{Classical Time-Series Model: ARIMA}

Autoregressive Integrated Moving Average (ARIMA) models have been widely adopted as baseline approaches for blood glucose level (BGL) forecasting due to their mathematical transparency, computational efficiency, and interpretability~\cite{s21051647}. However, the foundational assumptions of ARIMA, such as linearity and stationarity, are poorly matched to the nonlinear and time-varying dynamics of glucose regulation, which are shaped by multifactorial influences including insulin administration, carbohydrate intake, and physical activity. Therefore, ARX models extend the ARIMA framework by incorporating exogenous variables~\cite{ROMEROUGALDE2019321}. ARX model improves prediction accuracy during and after physical activity by integrating insulin dosage, carbohydrate intake, and energy expenditure. However, a major drawback of statistical and conventional machine learning approaches is their dependence on engineered features derived from the input data~\cite{georga2012multivariate}. This limitation is addressed by deep learning models. Deep learning methods, extensively used in fields such as distributed systems~\cite{alsadat2024distributed,alsadat2024using}, healthcare~\cite{mamun2025glucolens,azghan2025cudle}, and classification tasks~\cite{chatrudi2024wavelet,elhambakhsh2025domain}, enhance forecasting performance by capturing complex temporal relationships and nonlinear patterns in time series data.




\subsection{Shallow Neural Networks\looseness=-1}

Shallow feedforward neural networks, particularly multi-layer perceptrons (MLPs), represent some of the earliest neural architectures adopted for BGL prediction~\cite{naresh2024non,butt2021machine}. These models typically operate over fixed-length input windows and predict future glucose values using a static input-output mapping, making them suitable for real-time embedded deployment due to their low computational overhead and minimal latency~\cite{s21051647}. MLPs are fundamentally limited in their capacity to model complex temporal dependencies or nonlinear feedback inherent in physiological systems. Furthermore, MLPs are often integrated into hybrid frameworks, where they serve as denoising stages or feature embedding layers within more complex temporal architectures such as LSTMs or Transformers~\cite{Dweekat2023}. However, shallow network models still lack the ability to capture long-term complex temporal dependencies.



\subsection{Recurrent Neural Networks\looseness=-1}

Recurrent neural networks (RNNs), particularly Long Short-Term Memory (LSTM) and Gated Recurrent Unit (GRU) models, are widely used in BGL forecasting for their ability to capture medium-range temporal dependencies~\cite{9007528,li2019convolutional}. Their gating mechanisms mitigate vanishing gradient issues and allow the modeling of sequential physiological signals such as insulin action, meal timing, and glucose response. Thus,  LSTM and GRU architectures remain foundational baselines in glucose forecasting due to their strong short-term accuracy and capacity for multimodal integration~\cite{arefeen2023glysim}. Shuvo \textit{et. al} proposed a multitask deep learning model based on stacked LSTM architecture for personalized blood glucose forecasting, utilizing a combination of shared and individual-specific layers to capture both common patterns and personalized glucose behaviors~\cite{shuvo2023deep}. However, recent deep learning models such as Transformer can achieve better performance of forecasting time series data compared with the LSTM model due to the attention mechanism~\cite{vaswani2017attention,kim2024comprehensive}


\subsection{Transformer-Based Models}

Transformer-based architectures have recently emerged as a powerful alternative for modeling BGL time series, primarily due to their self-attention mechanisms, which enable the capture of long-range temporal dependencies between multivariate data without relying on recurrence~\cite{gluformer2023, bgformer2024}.
Several Transformer variants have been tailored specifically for glucose forecasting. BGformer, an adaptation of the Informer model, introduced a dual-attention strategy comprising overlapping segment attention and channel-mixing attention to enhance feature extraction from CGM data~\cite{bgformer2024}. It outperformed eight benchmark models, including LSTM and GRU baselines, achieving notable reductions in RMSE and MAE. In parallel, Gluformer incorporated hierarchical attention mechanisms and uncertainty quantification modules, achieving consistent RMSE improvements of 1–2 mg/dL over LSTM-based models in short-term prediction horizons~\cite{gluformer2023}. Furthermore \cite{farahmand2024hybrid} proposed GlucoNet, a hybrid deep learning framework that integrates LSTM and Transformer architectures with variational mode decomposition (VMD) to improve blood glucose forecasting from multimodal health data . By applying knowledge distillation to reduce model complexity, GlucoNet achieved up to 60\% improvement in RMSE compared with the LSTM model.\looseness=-1

Some Transformer architectures were not originally intended for CGM applications; they were initially developed for forecasting time series data but can be adapted for CGM use. Informer model~\cite{zhou2021informer} is a well-known, efficient Transformer-based approach designed for long-sequence time series forecasting. It addresses the computational and memory limitations of traditional Transformers by utilizing a ProbSparse self-attention mechanism and attention distillation. The Temporal Fusion Transformer (TFT) represents another influential Transformer architecture originally designed for general multihorizon time-series forecasting~\cite{lim2021temporal}. TFT integrates static covariates and dynamic inputs through interpretable attention layers and gating mechanisms.
Transformers also offer structural advantages for multimodal modeling. Their capacity to process heterogeneous feature streams via parallel attention layers makes them well-suited for integrating behavioral and physiological signals such as insulin dosages, heart rate, physical activity, and sleep stages~\cite{farahmand2024hybrid,lim2021temporal}. While Gluformer focuses on univariate CGM inputs, it highlights multimodal extension as a key direction for future work~\cite{gluformer2023}.

Despite their strengths, Transformer-based models face notable limitations. Both Gluformer and BGformer report challenges related to data sparsity, irregular sampling intervals, and temporal misalignment between input modalities~\cite{gluformer2023, bgformer2024}.  The summary and limitations of related works are illustrated in Table~\ref {tab:summary}.

\begin{table*}[ht]
\centering
\caption{Summary of Related Methods, Limitations, and Their Application}
\label{tab:summary}
\begin{tabular}{p{1.5cm} p{1.7cm} p{3.8cm} p{5cm} p{2.2cm}}
\hline
Method & Examples & Key Features & Limitations & Application  \\
\hline
ARIMA & \cite{9007528} & Linear autoregression; baseline interpretability & Low RMSE in nonlinear regimes; Not able to combine irregular multi variant data  & BGL Forecasting \\ \hline
Shallow NN & \cite{9007528,s21051647} & Adaptive to nonlinear patterns; faster inference & Lacks temporal modeling; limited by input encoding & BGL Forecasting  \\ \hline
LSTM  & Glysim~\cite{arefeen2023glysim} & Medium-range memory via gated cells & Sensitive to overfitting; recursive error compounding & BGL Forecasting  \\ \hline
GRU & \cite{9007528} & Fewer parameters than LSTM; fast convergence & Less effective on long-sequence dynamics  & BGL Forecasting \\ \hline
Transformer & Gluformer~\cite{bgformer2024}, BGformer~\cite{gluformer2023} & Hierarchical attention,; Predicts sample points and probability distributions & Large data demands, architecture complexity sensitive to input alignment, Not able to combine irregular multi variant data & BGL forecasting  \\ \hline
Transformer & Informer~\cite{zhou2021informer}, TFT~\cite{lim2021temporal} & Self-attention distillation, Hierarchical attention, Explainable Attention, Predict Long-term PH& Large data demands, sensitive to input alignment, Not able to combine irregular multi variant data & Generic  Time Series Application\\
\hline
\end{tabular}
\end{table*}

\section{Proposed Method}
\begin{figure*}
    \centering
    \includegraphics[width=0.85\linewidth]{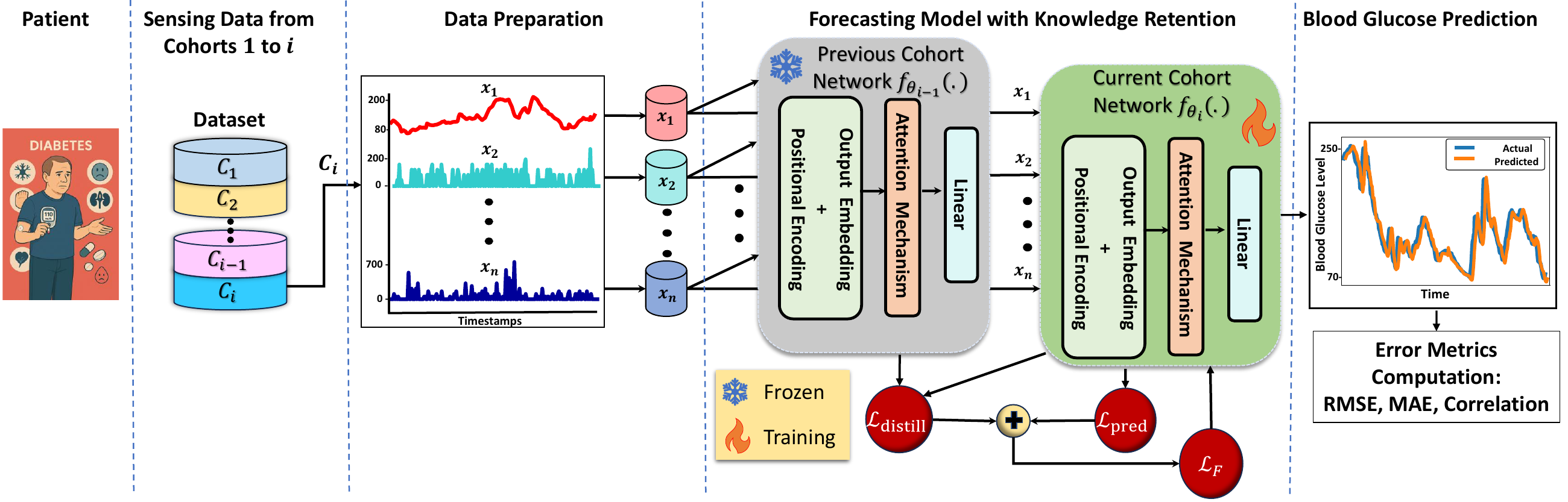}
    \caption{Overview of GluMind framework, which incorporates sensing data from the dataset and a Transformer-based forecasting model. The framework includes data preparation, the forecasting model with knowledge retention, and the blood glucose prediction.}
    \label{fig:overview}
\end{figure*}

In this section, we will elaborate on our proposed forecasting blood glucose level framework (GluMind) in detail. An overview of our proposed model is depicted in Fig.~\ref{fig:overview}. The framework is made up of four components: 1. Sensing module from various cohorts (e.g., cohorts 1 to $i$), that measures physiological and behavioral body signals from wearable sensors. These data are supplied by the dataset from different cohorts of diabetes patients. 2. Data preparation module, which preprocesses the time series data passed from the sensing module. 3. Forecasting model with Knowledge Retention, utilizing a novel machine learning forecasting model based on the Transformer architecture for blood glucose prediction by retaining knowledge from the previous cohort. This retention of knowledge from the previous cohort addresses the forecasting model's forgetting issue. 4. Blood Glucose Prediction, wherein the BGL prediction signal serves as the target of the forecasting model and the error metrics are computed. 
Our transformer-based model predicts BGL in individuals with type 2 diabetes by incorporating CGM data alongside physiological and body variables such as activity, heart rate, and stress data. The attention mechanism within the Transformer facilitates the effective fusion of multi-time series signals recorded at different sampling rates and captures their long-term temporal dependencies. Therefore, a Transformer-based forecasting model is well-suited for predicting highly fluctuating signals such as BGLs. To validate the effectiveness of our proposed model, we conduct experiments using the publicly available AI-READI (Flagship) dataset. The following sections provide a detailed explanation of the forecasting problem and key components of our proposed model.\looseness=-1
\subsection{Problem Formulation}
The multimodal blood glucose forecasting can be formulated as a time series prediction task in which \( \mathbf{X} = [\mathbf{x}_1, \mathbf{x}_2, \dots, \mathbf{x}_n] \) represent a set of \( n \) sensor-derived measurements in the sensing data component. The observation from the \( i\)th sensor is denoted as \( \mathbf{x}_i = [x_{i,1}, \dots, x_{i,t}]^\top \), where $t$ is the sampling duration. Assume that $\mathbf{x}_{\text{g}}$ is CGM data so that the forecasting BGL formulated by Eq.~\ref{eq:glucose_formulation}.
\begin{equation}
\mathbf{\hat{x}}_g = [x_{g,t+1}, \dots, x_{g,t+m}]^\top= f(\mathbf{X; \Theta})
\label{eq:glucose_formulation}
\end{equation}

\noindent where $m$ represents the number of predicted time steps, commonly referred to as the prediction horizon (PH). $f$ represents the forecasting model, which is developed using a Transformer architecture in this paper, parameterized by $\Theta$, which is learned during the training process.
 
\subsection{GluMind}

The two primary stages of GluMind framework are data preparation and forecasting model with knowledge retention. The data preparation stage focuses on collecting and processing physiological and behavioral data to serve as input for the forecasting model. In this paper, we used data collected by the AI-READI dataset. BGLs are measured using a CGM device, while other behavioral metrics, such as physical activity, stress, and heart rate, are recorded from wearable sensors such as smartwatches, as depicted in Fig.~\ref{fig:overview}. For processing data in this stage, the interpolation method is applied to time series data to handle missing values. Furthermore, the normalization is also applied to the data for consistency in the first stage. The second stage is the multimodal forecasting model with knowledge retention, which utilizes these preprocessed inputs for blood glucose prediction.\looseness=-1

The proposed forecasting model leverages a multimodal Transformer-based encoder architecture tailored for supervised learning, as depicted in Fig.~\ref{fig:model_architecture}. The vanilla Transformer architecture is commonly composed of an encoder-decoder structure designed for data reconstruction~\cite{vaswani2017attention}. However, in our forecasting model, the decoder is omitted to focus solely on latent representation learning. As shown in Fig.~\ref{fig:model_architecture}, our developed Transformer architecture includes two key attention mechanisms: cross-attention and multi-scale attention. The cross-attention mechanism facilitates the integration of multivariate time series data with varying sampling rates, while the multi-scale attention captures temporal dependencies within the signals and mitigates the impact of random noise~\cite{shabani2022scaleformer}. These enhancements contribute to improved accuracy in BGL forecasting.

In the AttenGluco study~\cite{farahmand2025attengluco}, the forecasting architecture employed a sequential configuration of cross-attention followed by a multiscale attention mechanism to model long-term dependencies and fuse irregularly sampled data. However, in our proposed model, we propose a parallel formulation of these attention mechanisms (Fig.~\ref{fig:model_architecture}). Deploying the attention mechanisms in parallel allows each to specialize in distinct representational learning tasks: the cross-attention module focuses on aligning and integrating physiological signals with BGL, while the multiscale attention module captures long-range temporal dependencies. This parallel attention design enhances the model's capability to represent complex temporal dynamics, resulting in better predictive performance.  
Furthermore, to address catastrophic forgetting (CF) across various cohorts, the knowledge retention method is applied to the forecasting model across multiple patient cohorts.\looseness=-1

Following sections will elaborate on the forecasting model’s architecture and its mechanisms for retaining knowledge from previous cohorts.\looseness=-1

\subsubsection{Forecasting Model Architecture}
The preprocessed data generated in the second stage of our framework (Fig.~\ref{fig:overview}) is passed to a Transformer-based forecasting model. This model takes the historical target data (e.g., CGM data), denoted as $\mathbf{x}_g$, along with other sensor-derived signals $\mathbf{x}_1, \dots, \mathbf{x}_n$ as input. In this paper, the CGM data serves as the prediction target, with other physiological variables incorporated as predictive features. Each input sequence is first mapped through an embedding layer $f_{\text{embed}}(\cdot)$, followed by a positional encoding function $f_{\text{pos}}(\cdot)$, yielding transformed representations $\mathbf{X}_g$ and $\mathbf{X}_1, \dots, \mathbf{X}_n$, respectively. Each resulting matrix lies in $\mathbb{R}^{t \times d_{\text{model}}}$, where $t$ denotes the time window length and $d_{\text{model}}$ is a model hyperparameter defining the embedding dimension. These representations are then processed in parallel by two distinct attention modules: a cross-attention mechanism and a multi-scale attention mechanism.\looseness=-1

The multi-head attention mechanism in vanilla Transformer architectures~\cite{vaswani2017attention} functions by scaling values \( (\mathbf{V} \in \mathbb{R}^{t \times d_\text{model}}) \) based on the relationships between keys \( (\mathbf{K} \in \mathbb{R}^{t \times d_{\text{model}}}) \) and queries \( (\mathbf{Q} \in \mathbb{R}^{t \times d_{\text{model}}}) \). The mathematical formulation of the attention mechanism is presented in Eq.~\ref{eq:attention}.

\begin{equation}
\text{Attention}(\mathbf{Q},\mathbf{K},\mathbf{V})= \text{Softmax} \left(\frac{\mathbf{Q} \mathbf{K}^T}{\sqrt{d_{\text{model}}}} \right)\mathbf{V}
\label{eq:attention}
\end{equation}

In a multi-head attention mechanism, it is essential to combine various physiological variables, each potentially sampled at different rates. Therefore, by conducting down-sampling or up-sampling to align all time series inputs with the target data (here is BGL). The embedded and position-encoded representations are subsequently concatenated into a unified sequence, which serves as the input to the multi-head attention module. This attention mechanism is capable of extracting semantic correlations across elements within long sequences~\cite{zeng2023transformers}.

\begin{figure}
    \centering
    \includegraphics[scale=0.2]{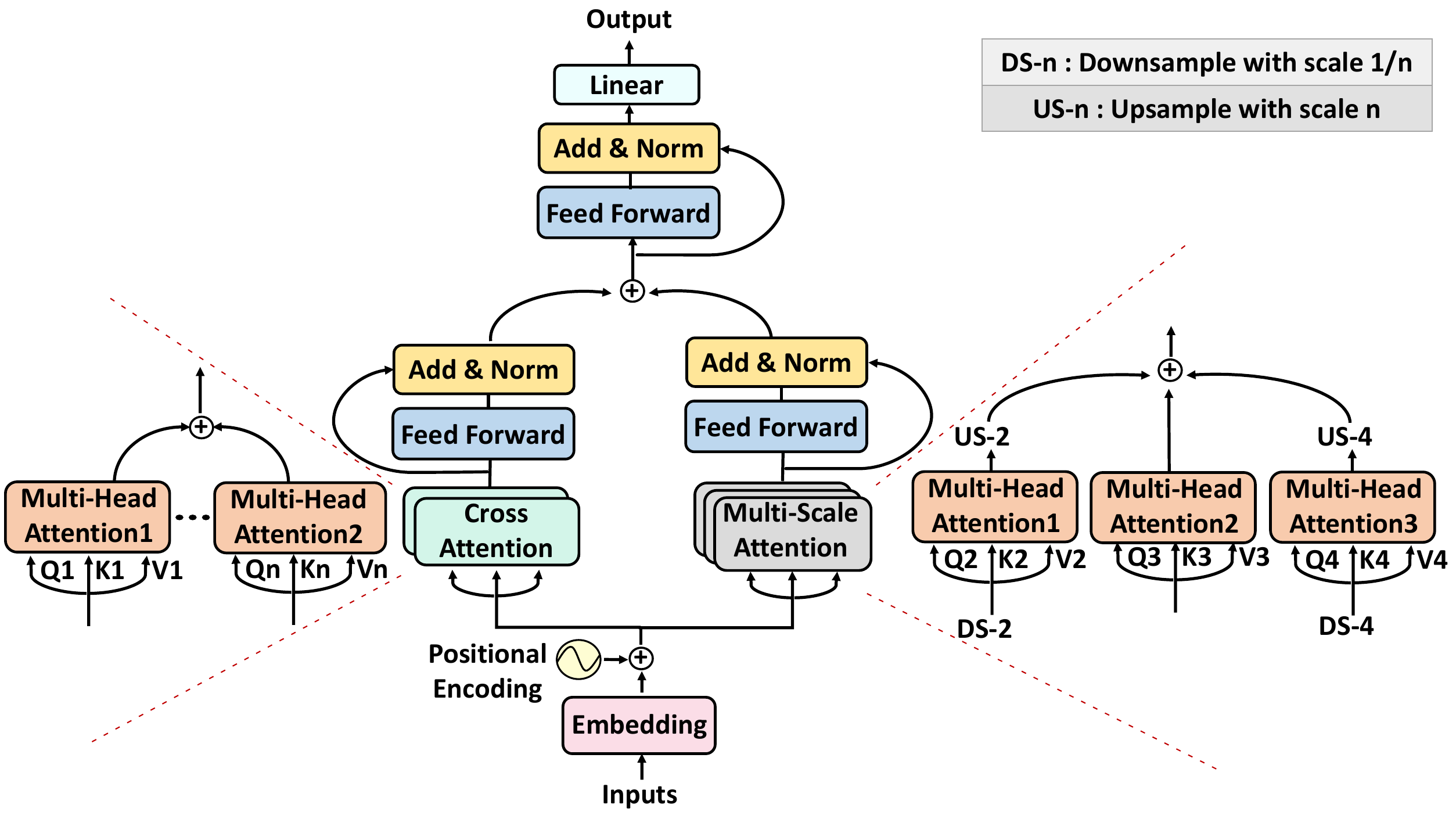}
    \caption{Our multimodal proposed model architecture for predicting BGLs, including Cross and Multi-scale attention.}
    \label{fig:model_architecture}
\end{figure}

To enhance the ability of multi-head attention to capture temporal dependencies, our forecasting model incorporates a multi-scale attention mechanism. The multi-scale attention framework addresses explicitly two key aspects: 1. Locality Perception, small-scale attention focuses on capturing local contextual dependencies, which are critical for understanding short-term dynamics in multivariate time series; 2. Long-term Hierarchical Representation, by combining hierarchical scales of attention, the model can leverage the inherent multi-scale temporal structures present in the data. Thus, combining various attention layers across different scales enables the model to capture both local context and higher-order global patterns~\cite{feng2023multi}.
The combined embedded and positional encoded inputs, \( \mathbf{X}_{\text{I}}=\{ \mathbf{X}_\text{G}, \mathbf{X}_\text{1}, ... \mathbf{X}_n\} \), are fed into a multi-scale attention mechanism comprising three multi-head attention branches, each designed for different downsampling (DS) rates. These branches apply downsampling factors of 1, 2, and 4, where a factor of 1 indicates no downsampling, as illustrated in Fig.~\ref{fig:model_architecture}.

For the first branch, the multi-scale attention mechanism (MA) on $\mathbf{X}_{\text{I}}$ is computed by using Eqs.~\ref{eq:multiatt} and~\ref{eq:multiatt2}.

\begin{equation}
        \text{MA}(\mathbf{X}_\text{I}, \mathbf{X}_\text{I}, \mathbf{X}_\text{I}) =
        [\mathbf{H}_1, \dots, \mathbf{H}_{m_H}] \mathbf{W}_H^{\text{MA}}
    \label{eq:multiatt}
\end{equation}
\begin{equation}
    \mathbf{H}_h = \text{Attention}(\mathbf{X}_\text{I} \mathbf{W}_{\mathbf{Q}}^{\text{MA}},\mathbf{X}_\text{I} \mathbf{W}_{\mathbf{K}}^{\text{MA}}, \mathbf{X}_\text{I} \mathbf{W}_{\mathbf{V}}^{\text{MA}})
    \label{eq:multiatt2}
\end{equation}

Each attention branch utilizes query, key, and value weight matrices, \( \mathbf{W}_{\mathbf{Q}}^{\text{MA}} \), \( \mathbf{W}_{\mathbf{K}}^{\text{MA}} \), and \( \mathbf{W}_{\mathbf{V}}^{\text{MA}} \), all belonging to \( \mathbb{R} ^{d_{\text{model}}\times d_{\text{model}}} \). The outputs from all attention heads are concatenated and projected back into the original model dimension using the final weight matrix \( \mathbf{W}_H^{\text{MA}} \in \mathbb{R}^{(m_H \cdot d_{\text{model}}) \times d_{\text{model}}} \). The remaining two branches follow the same computational process but operate on downsampled input data. This approach improves the model’s capability to capture both fine-grained details and long-term temporal dependencies within the input signals. Then, the outputs from the three multi-scale attention branches are summed and passed through a feed-forward network, an Add \& Norm block.

However, fusing data from variables with different sampling rates through down-sampling or up-sampling can lead to the loss of informative samples or the generation of additional, potentially unrelated samples. To address this challenge, we employ a cross-attention mechanism, which enables more effective alignment and integration of irregular time series inputs.
Cross-attention mechanisms have demonstrated significant success in fusing information from multiple modalities across various domains for useful downstream tasks~\cite{nagrani2021attention}. We can use from the beneficiary of cross attention to fuse different sensor-derived signals with different sample rates. Moreover, these data exchange information via the attention mechanism to compute the correlation between different predictive features on the prediction target. Thus, we design an $n$-branch cross-attention layer in which all branches share \( \mathbf{X}_\text{G} \) as the query input. In the $i$-th branch, the keys and values are derived from \( \mathbf{X}_{i} \). The cross-attention (CA) of the $i$-th branch is computed using Eqs.~\ref{eq:crossatt} and~\ref{eq:crossatt2}.

\begin{equation}
        \text{CA}\left(\mathbf{X_\text{G}}, \mathbf{X}_i, \mathbf{X}_i\right) \\
         =[\mathbf{H}_1, \dots, \mathbf{H}_{m_H}] \mathbf{W}_H^{\text{CA}}
    \label{eq:crossatt}
\end{equation}
\begin{equation}
    \mathbf{H}_h = \text{Attention}(\mathbf{\mathbf{X_\text{G}}} \mathbf{W}_{\mathbf{Q}}^{\text{CA}}, \mathbf{X}_i \mathbf{W}_{\mathbf{K}}^{\text{CA}}, \mathbf{X}_i \mathbf{W}_{\mathbf{V}}^{\text{CA}})
    \label{eq:crossatt2}
\end{equation}

\noindent where $\mathbf{W}_{\mathbf{Q}}^{\text{CA}}$, $\mathbf{W}_{\mathbf{K}}^{\text{CA}}$, and  $\mathbf{W}_{\mathbf{V}}^{\text{CA}}$ are weight matrices specific to the attention head and belong to $\mathbb{R} ^{d_{\text{model}}\times d_{\text{model}}}$. Moreover, $\mathbf{W}_H^{\text{CA}} \in \mathbb{R}^{(m_H \cdot d_{\text{model}}) \times d_{\text{model}}}$ is the final weight matrix that projects the concatenated attention head outputs into the original model dimension. The attention mechanisms for the remaining $n-1$ branches are computed separately by following the same procedure. Then, the attention outputs from all branches are added to incorporate cross-attention information. The resulting matrix is passed through a linear feedforward network, followed by an Add \& Norm module. 

The output of the multi-scale attention branch is summed with the output of the cross-attention branch and passed through a linear feedforward network, followed by an Add \& Norm module and a fully connected layer. This final configuration generates $m$ predicted target values. In this study, each BGL value prediction corresponds to a measurement taken every 5 minutes, meaning that $m$ samples collectively provide forecasts for $m\times5$ minutes into the future.

\begin{algorithm}
\small
\caption{GluMind Model with Knowledge Retention}
\label{alg:TransformerBGL}
\textbf{Input:} Preprocessed and normalized data, including target ($\mathbf{x}_{\text{g}}$) and $n$ auxiliary sensor-derived signals ($\mathbf{x}_1, \dots, \mathbf{x}_n$). Model parameters from the previous cohort $\theta_i$ (if available). \\
\textbf{Output:} Predicted BGL $\hat{\mathbf{x}}_g$ \\
\begin{algorithmic}[1]
\State \textbf{Begin}
\vspace{0.5em}
\State \hspace{0.2cm} $\mathbf{X}_g \gets f_{\text{pos}}(f_{\text{em}}(\mathbf{x}_g))$
\For{$i = 1$ \textbf{to} $n$}
\State \hspace{0.2cm} $\mathbf{X}_i \gets f_{\text{pos}}(f_{\text{em}}(\mathbf{x}_i))$
\EndFor
\vspace{0.5em}

\State \hspace{0.2cm} \textbf{// Cross-Attention Branch}
\vspace{0.3em}
\State \hspace{0.2cm} $\mathbf{X}_{\text{CA}} \gets 0$
\For{$i = 1$ \textbf{to} $n$}
    \State \hspace{0.6cm} $\mathbf{X}_{\text{CA}} \gets \mathbf{X}_{\text{CA}} + \text{CA}(\mathbf{X}_g, \mathbf{X}_i, \mathbf{X}_i)$
\EndFor
\State \hspace{0.2cm} $\mathbf{X}_{\text{CA}} \gets f_{\text{AN}}^{(1)}(f_{\text{FF}}^{(1)}(\mathbf{X}_{\text{CA}}))$
\vspace{0.5em}

\State \hspace{0.2cm} \textbf{// Multi-Scale Attention Branch}
\vspace{0.3em}
\State \hspace{0.2cm} $\mathbf{X}\gets [\mathbf{X}_G, \mathbf{X}_1, ...,\mathbf{X}_n]$
\State \hspace{0.2cm} $\mathbf{X}^{(2)}, \mathbf{X}^{(4)}\gets $ Downsample($\mathbf{X}$, 2), Downsample($\mathbf{X}$, 4)
\State \hspace{0.2cm} $\mathbf{X}_{\text{MS1}}\gets$ MS($\mathbf{X}, \mathbf{X}, \mathbf{X}$)
\State \hspace{0.2cm} $\mathbf{X}_{\text{MS2}}\gets$ Upsample(MS($\mathbf{X}^{(2)}, \mathbf{X}^{(2)}, \mathbf{X}^{(2)}$), 2)
\State \hspace{0.2cm} $\mathbf{X}_{\text{MS3}}\gets$ Upsample(MS($\mathbf{X}^{(4)}, \mathbf{X}^{(4)}, \mathbf{X}^{(2)}$), 4)
\State \hspace{0.2cm} $\mathbf{X}_\text{MS}\gets f_{\text{AN}}^{(2)}\left(f_{\text{FF}}^{(2)}\left(\mathbf{X}_\text{MS1}+\mathbf{X}_\text{MS2}+\mathbf{X}_\text{MS3}\right)\right)$
\vspace{0.5em}
\State \hspace{0.2cm} \textbf{// Final Prediction}
\vspace{0.3em}
\State \hspace{0.2cm} $\hat{\mathbf{x}}_g \gets f_{\text{lin}}(\mathbf{X}_{\text{CA}} + \mathbf{X}_{\text{MS}})$

\vspace{0.5em}
\State \hspace{0.2cm} \textbf{// LwF-based Knowledge Retention}
    \State \hspace{0.2cm} $\mathcal{L}_{\text{pred}} \gets \text{MSE}(\hat{\mathbf{x}}_g, \mathbf{y})$
\If{model trained on previous cohort exists}
    
\State \hspace{0.2cm} $\mathcal{L}_{\text{distill}} \gets \text{MSE}(f_{\theta_{i+1}}(\mathbf{x}), f_{\theta_i}(\mathbf{x}))$
\State \hspace{0.2cm} $\mathcal{L} \gets \mathcal{L}_{\text{pred}} + \lambda \cdot \mathcal{L}_{\text{distill}}$
\State \hspace{0.2cm} \text{Update model parameters via backpropagation on } $\mathcal{L}$
\Else
    \State \hspace{0.2cm} \text{Update model parameters via backpropagation on } $\mathcal{L}_{\text{pred}}$
\EndIf

\State \hspace{0.2cm} \textbf{return} $\hat{\mathbf{x}}_g$
\State \textbf{End}
\end{algorithmic}
\vspace{-0.8mm}
\end{algorithm}

\subsubsection{Knowledge Retention Method}

The integration of the aforementioned components enhances the model’s ability to generalize to new subjects' data as training advances. Nonetheless, a key property of any intelligent system is to behave as if it were exposed to all the data at once. Without this capability, the model tends to gradually forget knowledge acquired from previous tasks as it encounters new data. In the context of blood glucose forecasting, patients with diabetes may present varying severities of illness, and a model trained across cohorts with different characteristics may lose the ability to accurately predict for subjects it had previously seen. This phenomenon, known as catastrophic forgetting (CF), highlights the necessity of mechanisms that allow the model to retain prior knowledge while still adapting to new data.\looseness=-1

To mitigate forgetting and promote knowledge retention, we adopt a distillation-based training strategy inspired by Learning without Forgetting (LwF)~\cite{li2017learningforgetting}. Specifically, after completing the training on a cohort (task) $C_{i-1}$, we save a snapshot of the model’s parameters, denoted by $\theta_i$. When training on a new cohort (task) $C_i$, in addition to minimizing the standard prediction loss $\mathcal{L}_{\text{pred}}$ (e.g., mean squared error) on the current cohort, we introduce a distillation loss $\mathcal{L}_{\text{distill}}$ to enforce consistency between the outputs of the current model $\theta_{i}$ and the frozen model $\theta_{i-1}$. Mathematically, the total loss ($\mathcal{L_F}$) is computed as Eq.~\ref{eq:loss}.  

\begin{equation}
\mathcal{L_F} = \mathcal{L}_{\text{pred}}(f_{\theta_{i}}(\mathbf{x}), y) + \lambda \mathcal{L}_{\text{distill}}(f_{\theta_{i}}(\mathbf{x}), f_{\theta_{i-1}}(\mathbf{x}))
\label{eq:loss}
\end{equation}

where $\mathbf{x}$ is input from the new cohort, $y$ is the corresponding label (prediction values), and $\lambda$ controls the trade-off between prediction accuracy and knowledge preservation. Here, $\mathcal{L}_{\text{distill}}$  measures the discrepancy using mean squared error between the outputs of the current and previous models when evaluated on the same input $\mathbf{x}$.

In summary, Algorithm~\ref{alg:TransformerBGL} describes the data processing pipeline in GluMind.

\section{Results \& Discussion}
In this section, we first introduce the AI-READI dataset and perform experiments on this dataset. We then present the effectiveness of our proposed model, GluMind, by comparing it with other state-of-the-art prediction models. The comparison is carried out using standard evaluation metrics such as Root Mean Square Error (RMSE), Mean Absolute Error (MAE), and correlation analysis. The experimental details will be discussed in Section~\ref{Sec:Experiment_detail}.

\begin{table*}
\centering
\caption{Investigation of RMSE values for different predictive features across cohorts in the AI-READI dataset. Best scores are highlighted in bold.}
\label{Tab:RMSE_effect_variables}
\begin{adjustbox}{max width=\textwidth}
\begin{tabular}{lccccccccc}
\toprule
\textbf{Cohort} 
& \multicolumn{7}{c}{\textbf{RMSE}}  \\
\cmidrule(lr){2-8}
 & BG & BG+W & BG+Stress & BG+W+R & BG+W+Stress & BG+W+HR & BG+W+Stress+HR\\
\midrule
Healthy 
 & 18.39 & 15.34 & 17.15 & 15.81 & 15.45 & 15.39 & \textbf{15}\\
Pre-T2DM 
 & 19.63 & 16.74 & 18.22 & 17.31 & 15.98 & 15.95 & \textbf{15.75} \\
Oral 
 & 24.28 & 19.65 & 22.40 & 20.61 & 19.79 & 19.30 & \textbf{19.23} \\
Insulin 
 & 28.35 & 23.87 & 26.21 & 24.63 & 23.16 & 23.27 & \textbf{22.78} \\
\bottomrule
\end{tabular}
\end{adjustbox}
\end{table*}

\begin{table*}
\centering
\caption{Investigation of MAE values for different predictive features across cohorts in the AI-READI dataset. Best scores are highlighted in bold.}
\label{Tab:MAE_effect_variables}
\begin{adjustbox}{max width=\textwidth}
\begin{tabular}{lccccccccc}
\toprule
\textbf{Cohort}  
& \multicolumn{7}{c}{\textbf{MAE}} \\
\cmidrule(lr){2-8}
 & BG & BG+W & BG+Stress & BG+W+R & BG+W+Stress & BG+W+HR & BG+W+Stress+HR \\
\midrule
Healthy 
 & 13.24 & 10.74 & 12.21 & 11.08 & 10.88 & 10.80 & \textbf{10.58} \\
Pre-T2DM 
 & 14.21 & 11.72 & 12.96 & 12.13 & 11.29 & 11.20 & \textbf{11.08} \\
Oral 
 & 17.76 & 13.85 & 16.25 & 14.48 & 14.11 & 13.80 & \textbf{13.74} \\
Insulin 
 & 20.73 & 16.93 & 18.35 & 17.33 & 16.67 & 16.73 & \textbf{16.41} \\
\bottomrule
\end{tabular}
\end{adjustbox}
\end{table*}

\subsection{Dataset}
The dataset utilized in this study is the publicly available AI-READI Flagship Dataset, which aims to promote research in artificial intelligence and machine learning related to Type 2 Diabetes Mellitus (T2DM). It was collected from 1,067 participants across three sites in the United States and includes individuals both with and without T2DM, ensuring a balanced representation across sex, race, and diabetes severity. The dataset comprises four categories such as healthy individuals, those with prediabetes, individuals with T2DM who are on oral medication, and individuals with T2DM who are on insulin.

A key feature of the dataset is its multi-modal structure, where participants were monitored over ten days using a Dexcom G6 CGM for real-time blood glucose, a Garmin Vivosmart 5 for physical activity, stress, and heart rate variability, and a LeeLab Anura sensor for environmental factors such as air quality and temperature. The dataset also includes survey data, clinical assessments, and retinal imaging. Daily step counts are recorded via an accelerometer, with occasional gaps due to device recharging. The heart rate sensor also computed a stress index (0-100) based on heart rate variability.\looseness=-1

In this paper, CGM data and other body variables that affect BGL, such as stress, heart rate, physical activity signals, including walking activity and running activity (steps and intervals), are extracted as key features. After filtering out subjects with missing data, 896 participants are included in the final analysis, distributed as follows: 323 healthy individuals, 207 pre-T2DM, 258 with T2DM on oral medication, and 108 with T2DM on insulin.
\subsection{Experimental Details}
\label{Sec:Experiment_detail}
We evaluate the BGL forecasting performance of GluMind in comparison with Glysim~\cite{arefeen2023glysim}, our prior work AttenGluco~\cite{farahmand2025attengluco}, and Informer~\cite{zhou2021informer}, to demonstrate its improved accuracy and effectiveness. The Glysim proposed a forecasting model that combines a multimodal 1D-CNN and LSTM architecture to predict BGL. Informer is a state-of-the-art Transformer-based model designed for time series forecasting. Additionally, AttenGluco has recently utilized a Transformer-based forecasting model to predict BGL. The performance of these state-of-the-art methods is evaluated on the AI-READI dataset and compared against GluMind. The comparison is conducted using error metrics, including Root Mean Square Error (RMSE) and Mean Absolute Error (MAE), as well as correlation analysis. 
To comprehensively evaluate the performance of the GluMind model, we further investigate the impact of various physiological variables on BGL forecasting, along with the effects of different knowledge retention methods to mitigate the CF issue. Moreover, we study the influence of varying data history lengths, prediction horizons, and the incorporation of a dual attention mechanism to assess the robustness and effectiveness of our proposed method.

\subsubsection{Experimental setup}
In this paper, all forecasting models, such as Glysim, Informer, and AttenGluco, receive a sliding window of data history as input (e.g., 400 minutes). Training is conducted for 500 epochs with a learning rate of 0.001, optimizing the Mean Squared Error (MSE) using the AdamW optimizer. Forecasting performance is assessed across three prediction horizons (PHs): 5 minutes (1 next sample point), 30 minutes (6 next sample points), and 60 minutes (12 next sample points). To ensure consistency, each model undergoes five independent training runs. Model performance is assessed across all subjects, with comparisons based on RMSE~\cite{arefeen2023glysim}, MAE~\cite{arefeen2023glysim}, and Correlation~\cite{zhang2023joint}. Furthermore, the training and testing scenario in this study follows a fine-tuning approach, where the model is initially trained on one subject and subsequently fine-tuned sequentially across the remaining subjects, with each subject serving as both training and testing data. The fine-tuned model is then transferred to the subjects in the next cohort to continue the adaptation process.

Note that for evaluating the contribution of different physiological signals, the model is trained without applying a knowledge retention method. However, in other experiments, a knowledge retention method is incorporated during training, not only solving the CF issue but also leading to improvements in the model’s error metrics for blood glucose levels.

\subsubsection{Contribution of Different Physiological Variables}
We investigate the effect of different physiological variables (predictive features) on BGL forecasting (prediction target). The investigation of feature importance begins by first assessing the model's predictive performance using only CGM data (BG), subsequently progressively incorporating additional features such as activity, including walking (W) and running (R), stress, and heart rate (HR) to evaluate their impact on forecasting BGL. Tables~\ref{Tab:RMSE_effect_variables} and \ref{Tab:MAE_effect_variables} present the RMSE and MAE of different cohorts of the AI-READI dataset from seven ablation study experiments focused on the effect of predictive features on the prediction target. Four predictive features are found to significantly influence on BGL forecast in both error metrics: CGM data (BG), walking data including walking steps (WS) and walking interval (WI), stress, and HR. We conclude that various combinations of other features yield minimal or no performance gains and can lead to overfitting in some cases. Therefore, we will use a combination of CGM data (BG), walking data (including walking steps WS and walking interval WI), stress, and HR for BGL forecasting in the following experiments.


\subsubsection{Performance on Long-term BGL Forecasting}
\begin{figure*}
    \centering
    \includegraphics[scale=0.7]{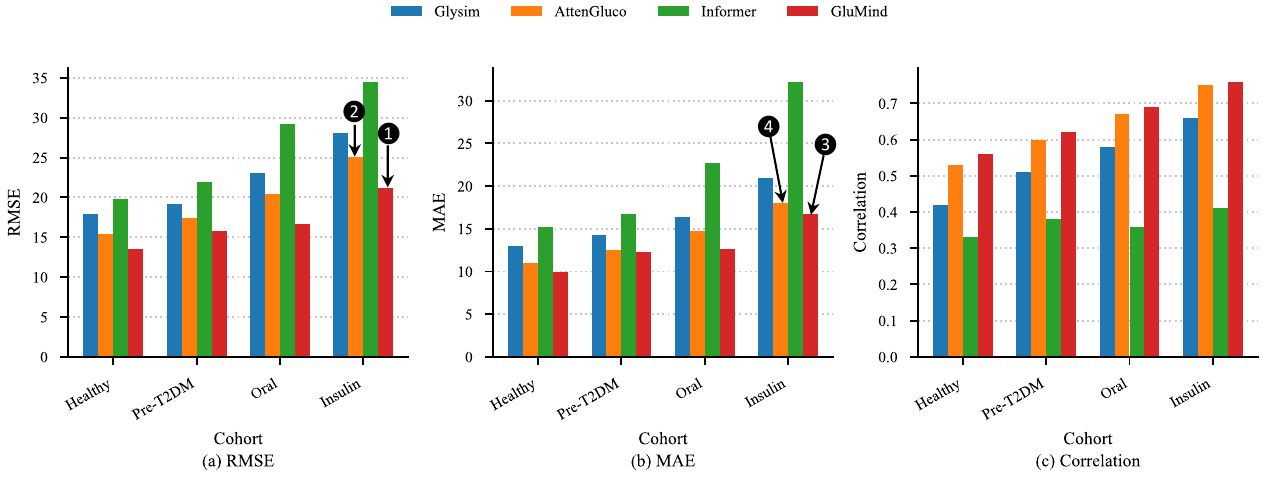}
    \caption{Comparison of GluMind model in terms of (a) RMSE, (b) MAE, and (c) Correlation across cohorts with the state-of-the-art forecasting models.}
    \label{fig:compareSOTA}
\end{figure*}

The comparison of our proposed model with other state-of-the-art forecasting models in terms of error metrics such as RMSE, MAE, and correlation is presented in Fig.~\ref{fig:compareSOTA}.
Fig.~\ref{fig:compareSOTA} illustrates that the GluMind model significantly outperforms other state-of-the-art models across all cohorts in the AI-READI dataset. For example, the RMSE of GluMind in the Insulin cohort is 21.17 (see pointer \textcircled{1}), while that of AttenGluco is 25.04 (see pointer \textcircled{2}), indicating an improvement of approximately 15\%. In terms of MAE, GluMind achieves 16.76 (see pointer \textcircled{3}), whereas AttenGluco achieves 18.03 (see pointer \textcircled{4}), demonstrating an improvement of around 9\% with our proposed model.



\begin{figure}
    \centering
    \includegraphics[width=\linewidth]{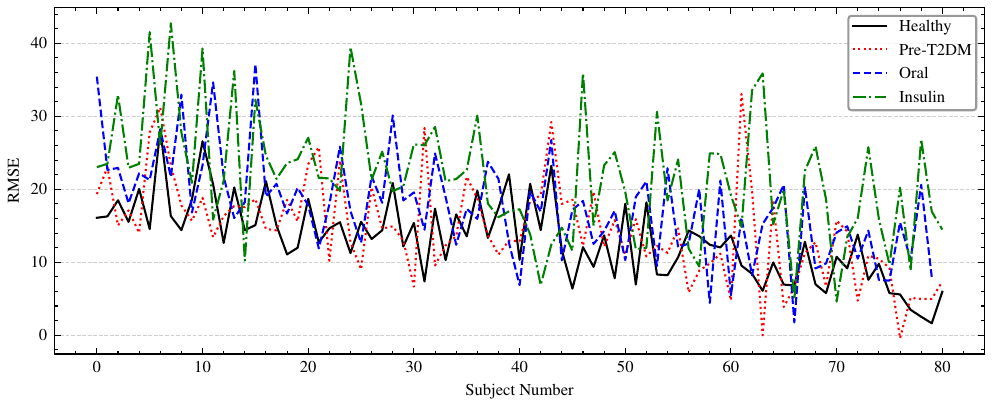}
    \caption{The RMSE of each subject in GluMind across different cohorts.}
    \label{fig:trendline}
\end{figure}

Moreover, Fig.~\ref{fig:trendline} demonstrates that as more subjects are added into each cohort, the model's performance progressively improves. Notably, the reduction in test error is more significant in our proposed model, indicating that its performance could further improve with a larger training dataset. For improved clarity and better visibility, we illustrate only 80 subjects of each cohort in Fig.~\ref{fig:trendline} while maintaining the overall distribution and trends of the complete dataset.
\subsection{Ablation Studies}
In this section, we perform a comprehensive series of ablation studies to evaluate the impact of each proposed architectural component and investigate various knowledge retention methods to address the CF issue. The outcomes of the different knowledge retention models are presented in the following section. Furthermore, we vary the data history and prediction horizon (PH), and systematically remove individual components from the model architecture, comparing the results to the original model.
\subsubsection{Effect of Knowledge Retention Method}

We used a distillation-based training, inspired by Learning without Forgetting (LwF)~\cite{li2017learningforgetting},  as our main knowledge retention approach to mitigate the effect of CF. During our experiments, we employed two alternative methods of knowledge retention to compare their effects and choose the best one. Namely, we used Elastic Weight Consolidation (EWC)~\cite{Kirkpatrick_2017}, and Experience Replay (ER)~\cite{rolnick2019experiencereplaycontinuallearning} to overcome the forgetting effect.
Table \ref{tab:clmethods} showcases the results we obtained using each method. In this table we report the following metrics:
\begin{itemize}
    \item Forgetting ratio: The ratio of model performance on a previously learned cohort after training on subsequent cohorts to its initial performance. This is formally defined as 
    \begin{equation}
    \text{FR}= \frac{\text{RMSE}_{\text{final}}}{\text{RMSE}_{\text{initial}}}
    \end{equation}
 where values greater than 1 indicate forgetting has occurred.
    \item Absolute forgetting: The absolute difference in error between initial and final performance on a task after training on subsequent tasks, which is calculated as\looseness=-1
    \begin{equation}
    \text{AF}= \text{RMSE}_{\text{final}} - \text{RMSE}_{\text{initial}}
    \end{equation}
    Higher positive values indicate more severe forgetting.
    \item Backward transfer: The percentage change in performance on previously learned tasks, quantifying knowledge transfer between tasks, which is computed as\looseness=-1
    \begin{equation}
    \text{BWT}= \frac{\text{RMSE}_{\text{final}} - \text{RMSE}_{\text{initial}}}{\text{RMSE}_{\text{initial}}} \times 100
    \end{equation}
    Negative values indicate positive backward transfer (improved performance), while positive values signify forgetting.\looseness=-1
\end{itemize}

\begin{table*}
\centering
\caption{Performance comparison of various knowledge retention algorithms integrated with our model across different cohorts in the AI-READI dataset.
}
\label{tab:clmethods}
\begin{tabular}{cccc}
\hline
            Knowledge Retention Method   &Avg. Forgetting Ratio $\downarrow$  & Avg. Absolute Forgetting $\downarrow$ & Avg. Backward Transfer $\downarrow$  \\\hline
None        & 1.1107                                        & 1.6430                                           & 11.07\%                                  \\
EWC & 1.0474                                        & 0.7493                                           & 4.74\%                                  \\
ER  &  1.0727 &                                            1.1575                                                  & 7.27\%    \\
 LwF (Used in GluMind) & \textbf{0.9249}                                        & \textbf{-0.2492}                                          & \textbf{-7.51}\%                                 \\

\hline                                   
\end{tabular}
\end{table*}

\begin{figure}
    \centering
    \includegraphics[width=\linewidth]{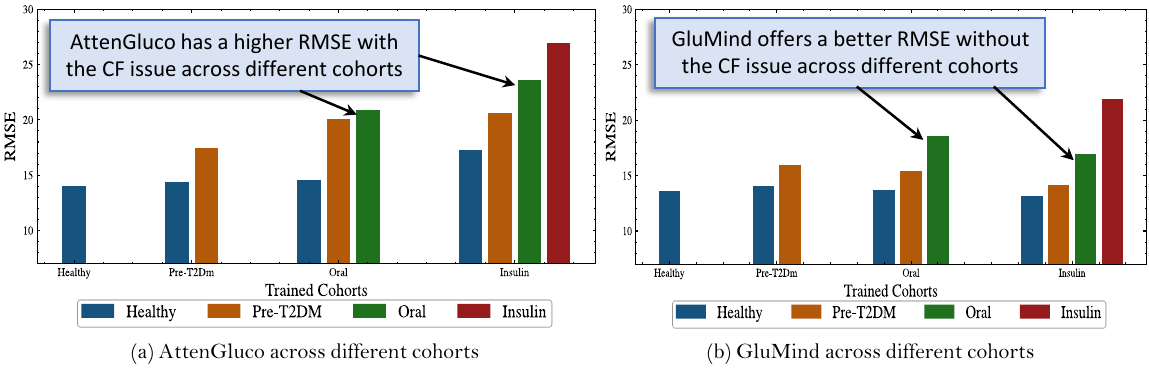}
    \caption{Comparing the AttenGluco algorithm with (b) the GluMind algorithm using LwF for mitigating CF across different cohorts of the AIREADI dataset.}
    \label{fig:CF mitigate}
\end{figure}

We can conclude that, compared to other methods, LwF not only mitigates the CF issue but also enhances the model’s accuracy as it encounters more subjects from various cohorts. The LwF method improves RMSE of different cohorts by about 8\%. Furthermore, Fig.~\ref{fig:CF mitigate} demonstrates both the accuracy improvement and the CF mitigation across different cohorts when using the LwF method. From Fig.~\ref{fig:CF mitigate}, it is evident that employing a knowledge retention method such as LwF improves the forecasting model’s accuracy and alleviates CF issues across cohorts.
\subsubsection{Varying Prediction Horizons and Input Data History}
We evaluate our proposed forecasting model's RMSE at different PH values of $5$, $30$, and $60$ minutes. Since CGM data is recorded at 5-minute intervals, a PH of 5 minutes corresponds to \( m = 1 \) sample, a PH of 30 minutes corresponds to \( m = 6 \) samples, and a PH of 60 minutes corresponds to \( m = 12 \) samples. Table~\ref{tab:PH_comparison} shows a comparison across different prediction horizons (PHs). GluMind achieved an average improvement of approximately 14.90\% over AttenGluco in 60-minute PHs across multiple cohorts. The RMSE of our proposed model remains relatively stable over different PHs, highlighting its robustness in long-term forecasting.\looseness=-1
\begin{table*}
\centering
\caption{RMSE values across different cohorts and prediction horizons (5 min, 30 min, and 60 min) for GlySim, AttenGluco, Informer, and our proposed model. Best-performing scores are highlighted in bold.}
\label{tab:PH_comparison}
\begin{tabular}{|c|ccc|ccc|ccc|ccc|}
\hline
\textbf{Cohort} & \multicolumn{3}{c|}{\textbf{GlySim}} & \multicolumn{3}{c|}{\textbf{AttenGluco}} & \multicolumn{3}{c|}{\textbf{Informer}} & \multicolumn{3}{c|}{\textbf{GluMind}} \\
\hline
 & 5 min & 30 min & 60 min & 5 min & 30 min & 60 min & 5 min & 30 min & 60 min & 5 min & 30 min & 60 min \\
\hline
Healthy & 7.35 & 14.37 & 17.79 & 7.63 & 12.38 & 15.45 & 15.42 & 17.33 & 19.81 & \textbf{4.26} & \textbf{9.83} & \textbf{13.56} \\
PreT2DM & 7.94 & 15.43 & 19.77 & 8.70 & 13.50 & 17.47 & 18.15 & 19.27 & 21.95 & \textbf{4.15} & \textbf{11.15} & \textbf{15.84} \\
Oral & 9.15 & 17.73 & 23.37 & 9.33 & 15.21 & 20.45 & 23.63 & 25.19 & 29.17 & \textbf{4.03} & \textbf{11.45} & \textbf{16.63} \\
Insulin & 12.11 & 21.00 & 28.22 & 11.94 & 18.55 & 25.04 & 28.54 & 30.12 & 34.51 & \textbf{5.83} & \textbf{14.62} & \textbf{21.17} \\
\hline
\end{tabular}
\end{table*}

\begin{figure}
    \centering
    \includegraphics[width=\linewidth]{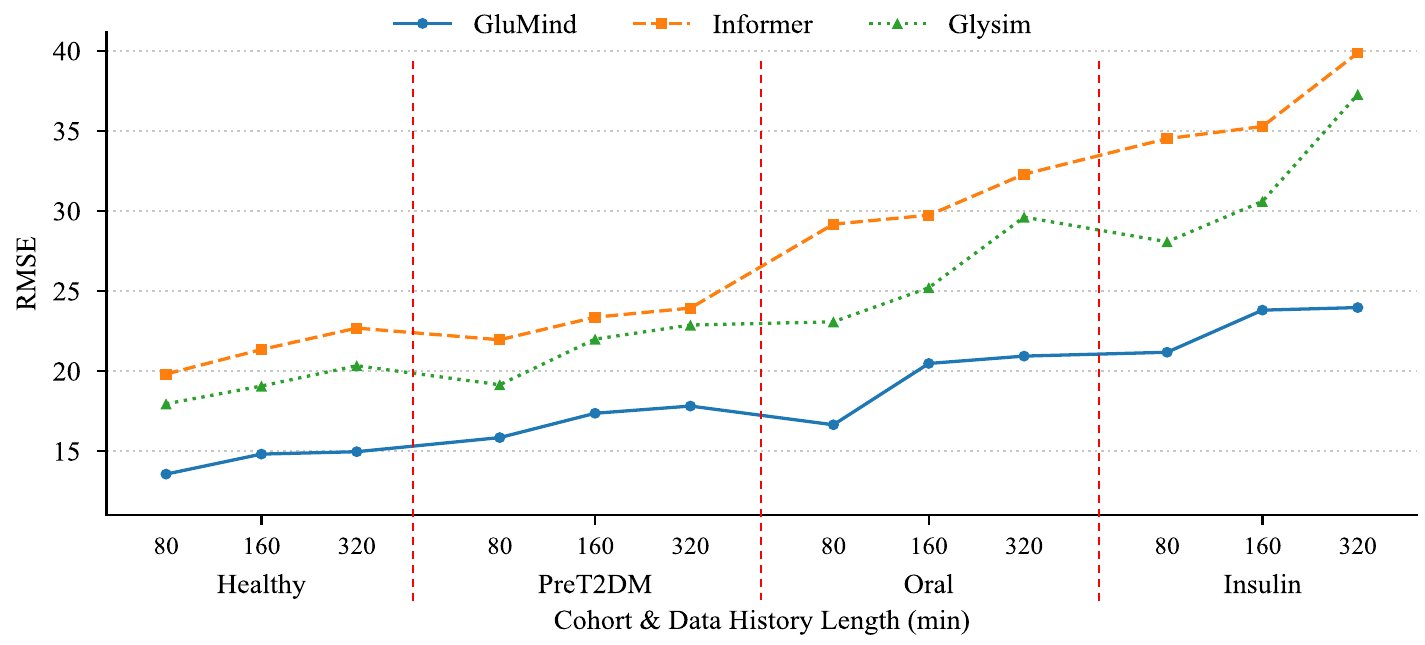}
    \caption{RMSE comparison across different cohorts and data history lengths (80 minutes (400 sample points)–320 minutes (1600 sample points)) for three models: GluMind, Informer, and Glysim. GluMind consistently achieves lower RMSE across all cohorts (Healthy, PreT2DM, Oral, Insulin), highlighting its robustness and generalization capability.}
    \label{fig:history of data}
\end{figure}

Moreover, we perform an ablation study using varying input data histories to evaluate the performance of the GluMind model. In principle, a longer data history expands the receptive field, which could potentially enhance forecasting accuracy. However, as discussed in~\cite{zeng2023transformers}, such improvements are not consistently observed in most forecasting models, particularly those based on transformers. Fig.~\ref{fig:history of data} illustrates the experimental ablation results using different input data histories across various cohorts from the AI-READI dataset. The scenario for this result is similar to the previous result, where training is transferred from Healthy cohorts to other cohorts. The results are presented in a line graph, segmented by a red dashed line to distinguish different cohorts and input history lengths. Fig.~\ref{fig:history of data} demonstrates that the GluMind model consistently achieves lower RMSE values as the receptive field increases, confirming its effectiveness in learning from extended input histories and its robustness and generalization capability.

\subsubsection{Impact of Dual Attention Mechanisms}
We conduct an ablation study on each attention mechanism by individually removing them from the model. Table~\ref{tab:ablateatten mech} presents the impact of these mechanisms on forecasting performance. Our model includes two attention mechanisms, such as cross and multi-scale attention, resulting in four different configurations for this study. We began by removing the cross-attention mechanism, leaving only the multi-scale attention in the model. The result of this configuration is reported as "Just Multi-Scale Attention" in Table~\ref{tab:ablateatten mech}. It can be observed that removing cross attention increases the RMSE of GluMind (which includes both attention mechanisms, as shown under "Cross and Multi-Scale Attention") from 21.17 to 25.45 in the Insulin cohort. Similarly, removing multi-scale attention leads to degraded forecasting performance. Furthermore, eliminating both attention mechanisms and replacing them with standard multi-head attention results in increased RMSE and MAE across all cohorts compared to the configuration with both attention mechanisms. The results demonstrate that both cross and multi-scale attention contribute significantly to overall model performance.

\begin{table*}
\centering
\caption{Comparison of RMSE and MAE across different cohorts under ablation study of the attention mechanism. The best metrics in terms of RMSE and MAE are highlighted in bold.}
\label{tab:ablateatten mech}
\begin{tabular}{|l|cc|cc|cc|cc|}
\hline
\textbf{Cohort} & \multicolumn{2}{c|}{\textbf{Just Multi-Scale Attention}} & \multicolumn{2}{c|}{\textbf{Just Cross Attention}} & \multicolumn{2}{c|}{\textbf{Multi-Head Attention}} & \multicolumn{2}{c|}{\textbf{Cross and Multi-Scale Attentions}} \\
                & RMSE & MAE & RMSE & MAE & RMSE & MAE& RMSE & MAE \\
\hline
Healthy         & 14.16 & 10.52 & 13.71 & 9.91 & 14.11 & 10.48 & \textbf{13.56} & \textbf{9.99} \\
PreT2DM         & 17.12 & 13.09 & 16.07 & 11.54 & 17.77 & 13.79 & \textbf{15.84} &\textbf{ 12.27} \\
Oral            & 20.03 & 15.00 & 18.98 & 13.22 & 21.73 & 16.44 & \textbf{16.64} & \textbf{12.59} \\
Insulin         & 25.45 & 19.19 & 22.64 & 15.98 & 26.21 & 19.95 & \textbf{21.17} & \textbf{16.76} \\
\hline
\end{tabular}
\end{table*}

\section{Conclusion}
In this study, we proposed GluMind, a Transformer-based multimodal framework designed to address key challenges in long-term blood glucose forecasting and mitigate CF issue. Enhanced predictive accuracy in GluMind is achieved by employing cross-attention and multi-scale attention mechanisms. These allow for the effective fusion of heterogeneous physiological signals with CGM data and the capture of significant long-range temporal dependencies. Moreover, to mitigate the CF issue during fine-tuning the model across various cohorts, we incorporated the knowledge retention method, such as the Learning without Forgetting (LwF) strategy, enabling the model to retain prior knowledge. We validated the effectiveness of GluMind using the AIREADI dataset across multiple diabetic cohorts. Experimental results demonstrate that our model consistently outperforms state-of-the-art forecasting models, with significant improvements in RMSE, MAE, and correlation metrics. Additionally, GluMind showed strong adaptability and performance stability as new cohorts were introduced, highlighting its robustness and generalization capabilities.
\section{Acknowledgment}
This work was supported in part by the National Science Foundation under grant IIS-2402650. Any opinions, findings, conclusions, or recommendations expressed in this material are those of the authors and do not necessarily reflect the views of the funding organizations.

\bibliographystyle{IEEEtran}
\bibliography{refs}

\end{document}